%% file: main.tex
\documentclass[letterpaper, 10 pt, conference]{ieeeconf}  % Comment this line out if you need a4paper

\IEEEoverridecommandlockouts                              % This command is only needed if 
\usepackage{times}
\usepackage{epsfig}
\usepackage{graphicx}
\usepackage{amsmath}
\usepackage{amssymb}
\usepackage{pifont}
\usepackage{multirow}
\usepackage{dsfont}
\usepackage{xspace}
\usepackage{cite}
\usepackage{xcolor}
\usepackage{tabulary}
\usepackage{mathtools}
\usepackage{booktabs}
\usepackage[normalem]{ulem}
\usepackage[caption=false]{subfig}
\usepackage[pagebackref=true,breaklinks=true,colorlinks,bookmarks=false]{hyperref}
\input{math_definition.tex}

\definecolor{green}{RGB}{51, 204, 51}
\definecolor{blue}{RGB}{0, 0, 255}
\definecolor{orange}{RGB}{255, 153, 51}

\makeatletter
\DeclareRobustCommand\onedot{\futurelet\@let@token\@onedot}
\def\@onedot{\ifx\@let@token.\else.\null\fi\xspace}
\def\eg{\emph{e.g}\onedot}

\def\ie{\emph{i.e}\onedot}

\makeatother

\useunder{\uline}{\ul}{}

\newcommand{\nets}{PLUMENet}

\title{\LARGE \bf
PLUMENet: Efficient 3D Object Detection from Stereo Images
}

\author{Yan Wang$^{3}$ \quad Bin Yang$^{1,2}$ \quad Rui Hu$^{1}$ \quad Ming Liang$^{1}$ \quad Raquel Urtasun$^{1,2}$ % <-this % stops a space
\thanks{$^{1}$Waabi.}%
\thanks{$^{2}$University of Toronto. Correspondence: \small\texttt{yw763@cornell.edu, byang@cs.toronto.edu, urtasun@cs.toronto.edu}}%
\thanks{$^{3}$Cornell University.}%
\thanks{$^{*}$This work was done by all authors while at Uber ATG.}%
}

\begin{document}

\maketitle

\input{1_abs.tex}
\input{2_intro.tex}
\input{3_relate.tex}
\input{4_method.tex}
\input{5_exp.tex}
\input{6_discuss.tex}

{\small
\bibliographystyle{IEEEtran}
\bibliography{main}
}

\input{7_supp.tex}

\end{document}

%% file: math_definition.tex
\usepackage{amssymb}
\usepackage{amsmath,amsfonts}
\usepackage{amsopn}
\usepackage{bm} % bold symbol
\usepackage{multirow}

\newcommand{\ProbOpr}[1]{\mathbb{#1}}
\newcommand{\expect}[2]{%
\ifthenelse{\equal{#2}{}}{\ProbOpr{E}_{#1}}
{\ifthenelse{\equal{#1}{}}{\ProbOpr{E}\left[#2\right]}{\ProbOpr{E}_{#1}\left[#2\right]}}} % Expectation: syntax: E{1}{2} = E_1[2], E{}{2}=E[2], E{1}{} = E_1
\newcommand{\var}[2]{%
\ifthenelse{\equal{#2}{}}{\ProbOpr{VAR}_{#1}}
{\ifthenelse{\equal{#1}{}}{\ProbOpr{VAR}\left[#2\right]}{\ProbOpr{VAR}_{#1}\left[#2\right]}}} % Expectation: syntax: V{1}{2} = V_1[2], V{}{2}=V[2], V{1}{} = V_1
\newcommand{\sL}{\mathcal{L}}

\newcommand{\eat}[1]{}
\newcommand{\method}[1]{\textsc{#1}}

\newcommand{\SRCNN}{\method{S-RCNN}\xspace}

\newcommand{\APBEV}{AP$_\text{BEV}$\xspace}

%% file: 1_abs.tex
%!TEX root=main.tex
\begin{abstract}
3D object detection is a key component of  many robotic applications such as self-driving vehicles.
While many approaches rely on expensive 3D sensors such as LiDAR to produce accurate 3D estimates, methods that exploit stereo cameras have recently shown promising results at a lower cost.
Existing approaches tackle this problem in two steps: first depth estimation from stereo images is performed to produce a pseudo LiDAR point cloud, which is then used as input to a 3D object detector.
However, this approach is suboptimal due to the representation mismatch, as the two tasks are optimized in two different metric spaces. 
In this paper we propose a model that unifies these two tasks and performs them in the same metric space.
Specifically, we directly construct a \textit{p}seudo \textit{L}iDAR feature vol\textit{ume} (\textit{PLUME}) in 3D space, which is then used to solve both depth estimation and object detection tasks.
Our approach achieves state-of-the-art performance with much faster inference times when compared to existing methods on the challenging KITTI benchmark~\cite{geiger2013vision}. 
\end{abstract}

%% file: 2_intro.tex
%!TEX root=main.tex
\section{Introduction}

Self-driving vehicles have the  potential to revolutionize the future of mobility.
In order to build a robust and safe autonomy system, reliable real-time perception is a necessity. Many self-driving cars are equipped with expensive LiDAR sensors, which provide accurate depth measurements that facilitate reasoning in  3D metric space. In recent years LiDAR based 3D detectors \cite{chen2017multi,yang2018pixor,qi2018frustum,lang2019pointpillars,shi2019pv} have shown very promising results. 
However, LiDAR sensors remain costly and have limited range, limiting their applicability.

\begin{figure}[t]
\begin{center}
	\includegraphics[width=1.0\linewidth]{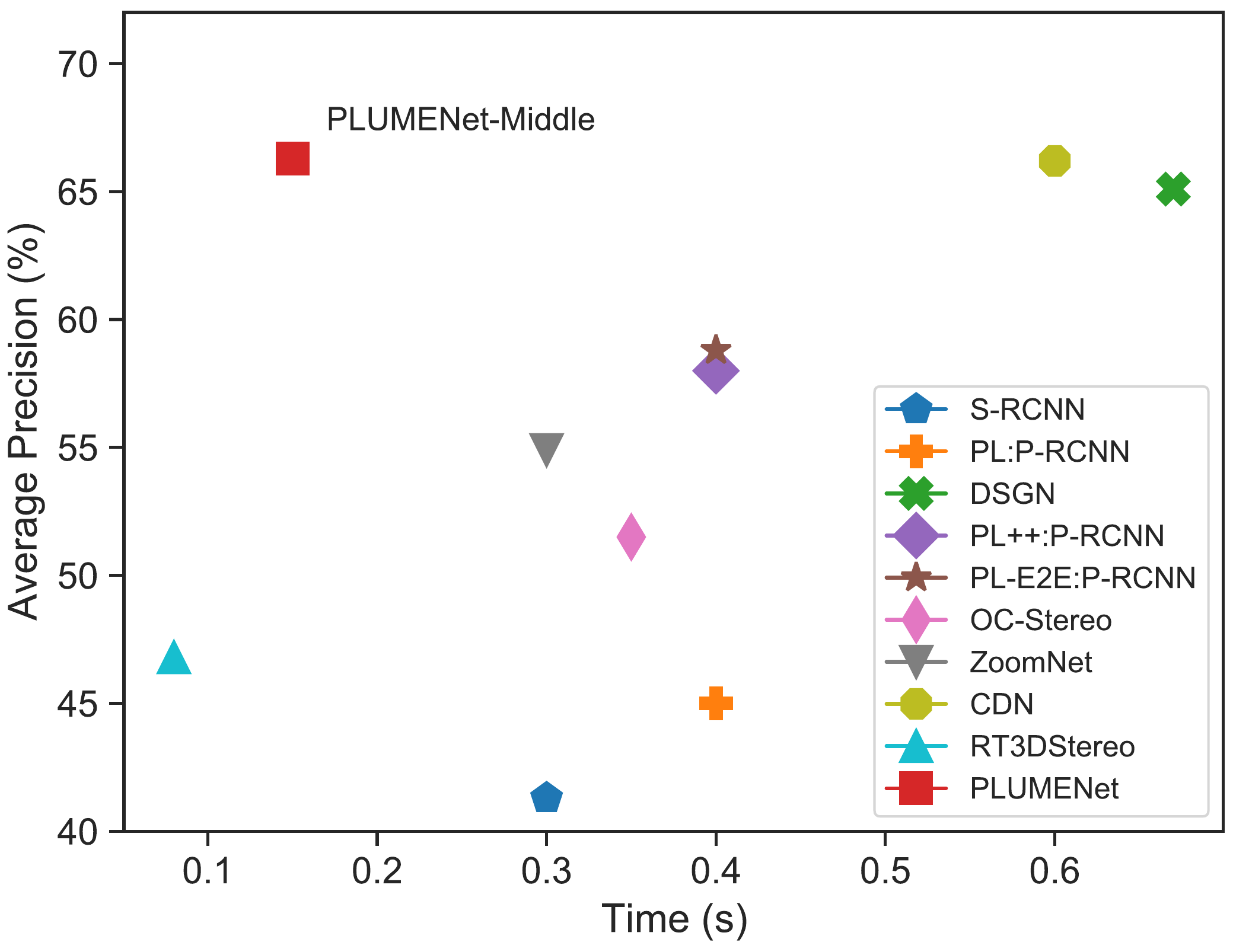}
\end{center}
\vspace{-3mm}
\caption{\textbf{Inference Time v.s. Average Precision} on KITTI~\cite{geiger2013vision} testing set. We report the average precision (AP) in bird's eye view (BEV) at moderate level. Our PLUMENet-Middle achieves state-of-the-art 66.3\% AP while running at 6.6 Hz.}
\label{fig:time_test}
\end{figure}

\begin{table*}[t]
\label{tbl:represent}
\begin{center}
\caption{Comparisons of different 3D feature volume (FV) representations.}
\begin{tabular}{@{}lllll@{}}
	\toprule
	& \textbf{Disparity FV} & \textbf{Depth FV} & \textbf{Pseudo-LiDAR FV} & \textbf{LiDAR FV} \\ \midrule
	\textbf{References}         &   \cite{chang2018pyramid,kendall2017end}                  & \cite{chen2020dsgn,you2019pseudo}                  & Ours                                 & \cite{yang2018pixor}                     \\
	\textbf{Data density}       & dense                          & dense                      & dense                                & sparse                        \\
	\textbf{Depth distribution} & distorted under perspective projection                    & distorted under perspective projection                & uniform                              & precise but sparse            \\
	\textbf{Metric space}       & image space                    & image space                & 3D/BEV space                         & 3D/BEV space                  \\
	\textbf{Feature extraction} & 3D convolution                 & 3D convolution             & 3D/2D convolution                    & 3D/2D convolution             \\ \bottomrule
\end{tabular}
\end{center}
\end{table*}

On the other hand, cameras are cheaper and much more accessible  than LiDAR. Even when LiDAR is available, cameras can improve the robustness of perception \cite{liang2019multi} as they provide high-resolution texture information. 
A major disadvantage of cameras is the difficulty of 3D reasoning.
Some approaches \cite{chen2016monocular, xu2018multi} use image features to regress 3D bounding boxes. However, as images are distorted representations of 3D information, reasoning in image space makes it hard to achieve precise 3D localization.
Pseudo-LiDAR-based approaches \cite{pseudoLiDAR} first estimate  depth from images and then transform the predicted depth image into 3D points (mimicking LiDAR), which can be processed by a standard 3D detector. The performance of pseudo-LiDAR directly depends on the quality of the estimated depth.  
Stereo images provide disparity cues that enable better depth estimation compared to monocular images. Recent stereo based pseudo-LiDAR models \cite{garg2020wasserstein,pseudoLiDAR,you2019pseudo} have greatly reduced the performance gap between camera and LiDAR based 3D detectors. These models usually construct the cost volume in image coordinate space with disparity or depth  as the third dimension. Matching costs are computed for all voxels and depth is then estimated by taking the local minimum along the third dimension (\ie, disparity or depth).

\begin{figure}[t]
\begin{center}
	\includegraphics[width=1.0\linewidth]{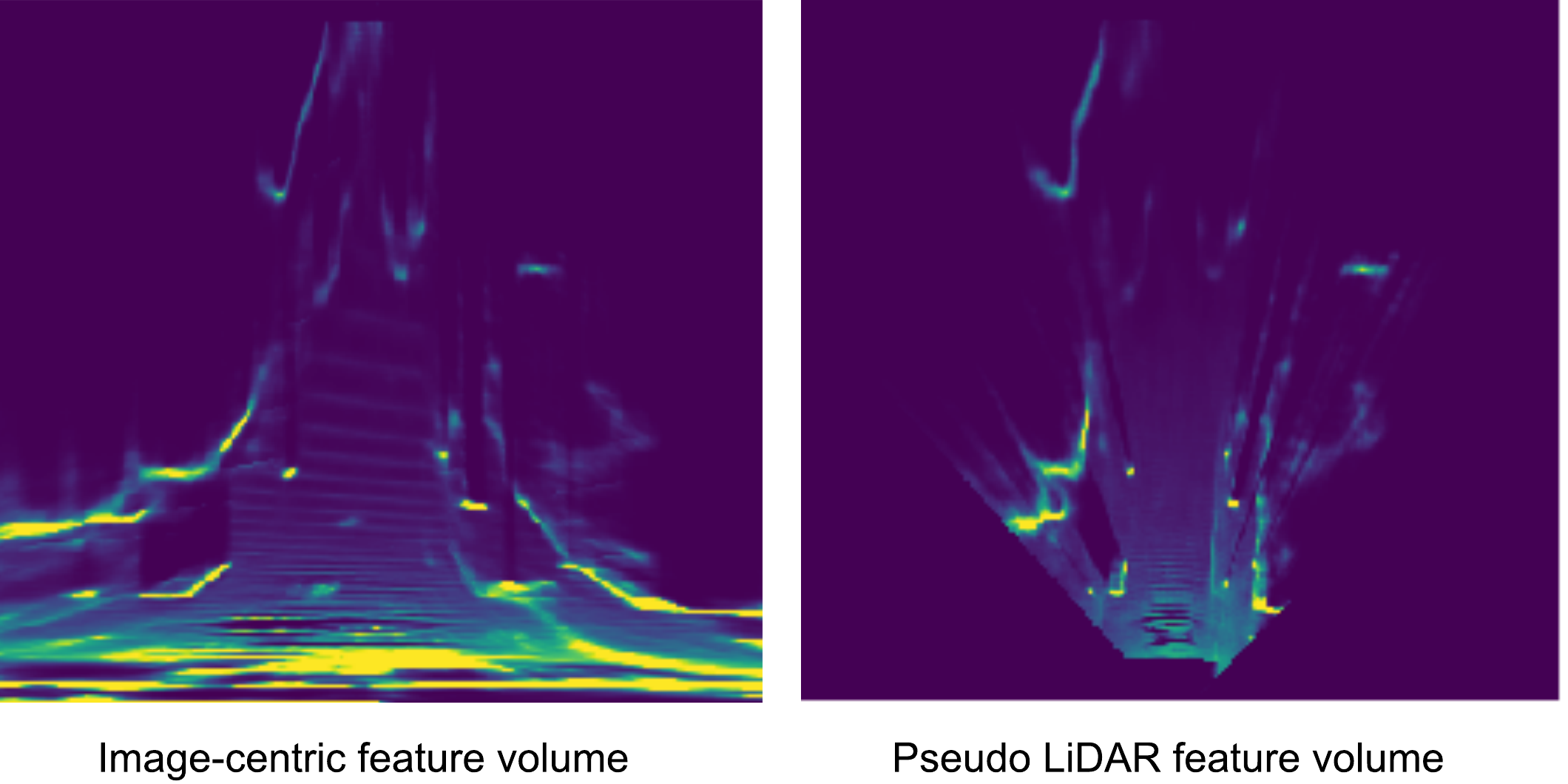}
\end{center}
\vspace{-3mm}
\caption{Feature volumes in image space and 3D metric space.}
\label{fig:cost}
\end{figure}

Although stereo based 3D detection models have shown promising results, they have several shortcomings. First, learning the 3D cost volume in image space is very costly in memory and computation due to the image's high resolution and the extra dimension required for the features. The typical processing time for one frame is 0.5 seconds \cite{li2020confidence,chen2020dsgn}, which is prohibitive in a safety critical application such as self-driving, where reaction time is key for collision avoidance. Second, depth estimation is performed in image space while the downstream detection is performed in 3D space. Due to perspective projection, nearby objects occupy many more pixels than faraway objects. This imbalance may lead to biased depth estimation with degraded long-range detection performance. This distortion effect is clearly shown in Fig. \ref{fig:cost}, where we visualize the feature volumes in image space and 3D metric space respectively.

To overcome the two aforementioned problems, in this paper we directly construct a \textit{p}seudo \textit{L}iDAR feature vol\textit{ume} (\textit{PLUME}) in 3D metric space from which we perform both depth estimation and 3D object detection.
As a result, we can build PLUME to leverage the full resolution image features. This contrasts with existing methods that have to heavily downsample image-centric features to keep memory and computation costs reasonable. 
Thus, our approach allows for greater efficiency while enabling superior detection performance.
Furthermore, with our formulation, 3D occupancy directly supervises the depth estimation model, thereby avoiding the imbalance issue that image-based metrics create.

We compare the effectiveness of our approach  on the challenging KITTI benchmark \cite{geiger2012we}. As shown in Fig. \ref{fig:time_test}, our model achieves 66.3\% average precision (AP) at moderate level on the testing set, surpassing all previously published methods that do not use external training data. Importantly, the inference time of our model is only 150 milliseconds, which is 4 $\times$ faster than other high-accuracy detectors \cite{garg2020wasserstein,chen2020dsgn}.

%% file: 3_relate.tex
%!TEX root=main.tex
\section{Related Work}

\begin{figure*}[t]
\begin{center}
    \includegraphics[width=\textwidth]{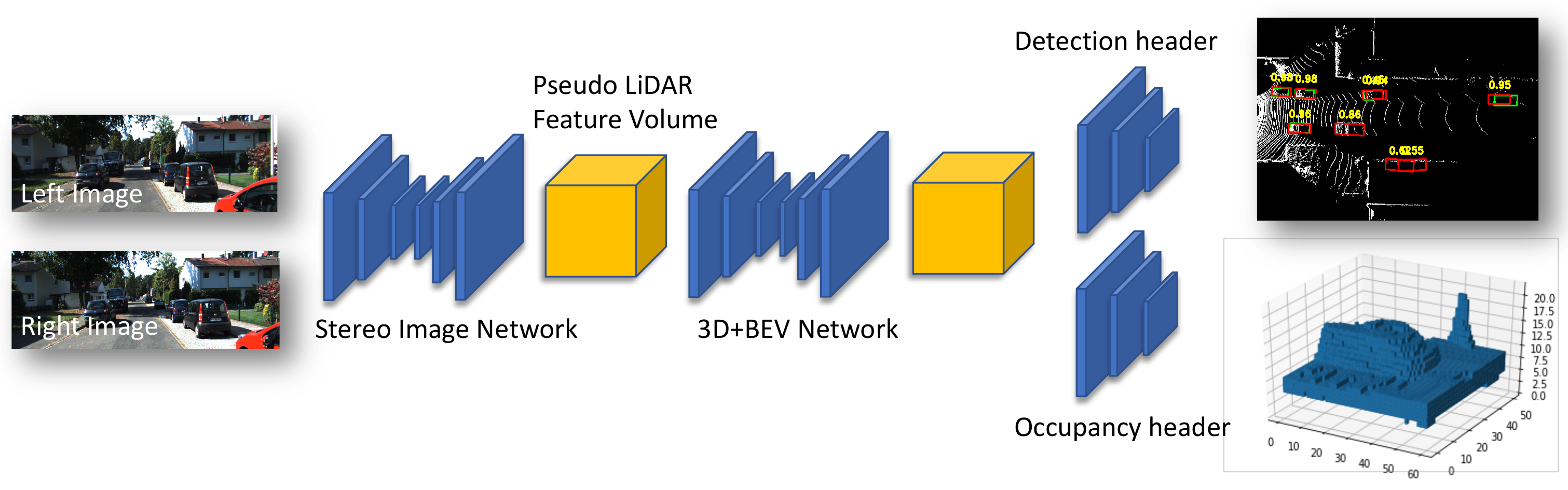}
\end{center}
\vspace{-3mm}
\caption{\textbf{Overview of \nets{}}: (1) stereo image network, (2) pseudo LiDAR feature volume (PLUME), (3) 3D-BEV network, (4)  multi-task headers. The network takes as input a stereo pair, and outputs the reconstructed 3D scene in the form of an occupancy grid and object bounding boxes in bird's eye view (BEV).}
\label{fig:overview}
\end{figure*}

\paragraph{Stereo Depth Estimation} 
Estimating depth from stereo images has been extensively investigated.
The fundamental principle is to create  correspondences between the left and right images. Existing depth estimation pipelines typically consists of four steps~\cite{scharstein2002taxonomy}: (1) Computing the matching cost of image or feature patches over a range of disparities; (2) Regularizing and smoothing the matching cost via aggregation; (3) Inspecting the low-cost regions to estimate  disparities; (4) Applying post-processing to refine the disparity results. 
Deep neural networks have been exploited as part of this pipeline.  \cite{zbontar2016stereo} uses a convolutional network to compute the matching cost  followed by non-learned based cost aggregation and disparity refinement. \cite{luo2016efficient} further improves the performance and efficiency of this approach by using an efficient Siamese network. \cite{kendall2017end} utilized a 3D disparity feature volume in image space, dimensions of which are image height $\times$ image width $\times$ disparity. Then a 3D convolutional network is applied to the feature volume for cost aggregation. Since the depth is inversely proportional to disparity, a small disparity represents a large depth. \cite{you2019pseudo} argues that using the disparity feature volume will over-optimize the close-range depth and introduces a depth feature volume that replaces the disparity with depth. This design change allows the model to directly output depth. \cite{garg2020wasserstein} proposes a new loss function based on the Wasserstein distance to further improve the depth prediction on  object boundaries.

\paragraph{LiDAR-based 3D Object Detection}
LiDAR sensors capture high quality depth measurements and facilitate many 3D reasoning tasks. LiDAR based 3D detectors  are extensively used by self-driving vehicles to perform accurate 3D perception. These methods can be divided into two categories, depending on the LiDAR point encoding utilized. Voxelization-based methods  \cite{chen2017multi, zhou2018voxelnet, yang2018pixor, yang2018hdnet, liang2018deep, liang2019multi} divide the 3D space into voxels, which are represented by learned or hand-crafted features on 3D points. 
Then 3D or bird's eye view (BEV) convolution can be applied to extract 3D features used by the detection header. The second category of methods utilize graph operators to directly process the points \cite{qi2017pointnet++, wang2018deep, shi2019pointrcnn}. The features and connections are directly computed from the point coordinates and the spatial proximity of points. Different from 3D and BEV convolution, point based operations make use of the sparsity of points and produce per-point output features. 
These two types of representations make the trade-off between efficient computation (with better locality of data structure) and precise information (without discretization error). Recently fusion based approaches \cite{shi2019pv, liu2019pvcnn} have been proposed to combine them together.

\paragraph{Image-based 3D Object Detection} 
Early works utilized a 2D frontal-view (monocular image) detection pipeline~\cite{he2017mask,lin2017feature,ren2015faster}, however, the 3D detection performance lags far behind  LiDAR-based solutions ~\cite{chabot2017deep,chen20153d,chen20183d,chen2016monocular,mousavian20173d,li2019gs3d,pham2017robust,xiang2015data,xiang2017subcategory,xu2018multi,roddick2019orthographic}. This gap has been significantly narrowed by  pseudo-LiDAR based approaches~\cite{pseudoLiDAR}, which  instead of directly predicting the 3D bounding boxes from monocular images, first  estimate depth at each pixel. Then the camera pose is used to un-project the pixels into 3D space to produce the pseudo-LiDAR representation. After that, standard LiDAR-based 3D detectors can be applied on the pseudo-LiDAR. OC-Stereo~\cite{pon2019object}, CG-stereo~\cite{li2020confidence} and disp-RCNN~\cite{sun2020disp} add auxiliary 2D bounding box regression or instance segmentation tasks to help distinguish the background from foreground pixels.
PL-E2E~\cite{qian2020end} proposed an end-to-end version of  this framework  where the detection loss supervises the depth network. 
However, as the core problem of representation mismatch is not addressed, the performance gain brought by end-to-end learning is minimal.
In contrast, in this paper we unify the representation utilized by the depth estimator and the object detector by employing a pseudo LiDAR feature volume. As a result, both tasks are optimized under the same 3D metric space, and most of the feature computation is shared, leading to a faster solution with superior detection accuracy.

Similar ideas of building 3D feature volumes directly from camera images has also been explored in previous works \cite{kar2017learning, roddick2019orthographic, chen2020dsgn}. 
Our work differs from them in the following aspects. 
While \cite{kar2017learning} only focuses on 3D object reconstruction, we extend it to the full scene and optimize the representation jointly with the downstream task of 3D object detection. 
\cite{roddick2019orthographic} builds the 3D feature volume from monocular images only and learns with 3D detection loss alone, while our approach exploits stereo matching cues to resolve the inherent depth ambiguity with auxiliary occupancy supervision. 
In \cite{chen2020dsgn} the same multi-task setting is explored with much worse performance, because it simply replaces the traditional image-centric feature volume as a 3D one without any network architecture change. In our work we find that traditional architecture is sub-optimal for the 3D feature volume. Specifically, the stereo image network needs to output high-resolution image features (instead of downsampled ones) so that the reconstructed 3D feature volume can capture fine details in stereo images to reason about depth especially in the long range. For the 3D convolutions used on 3D feature volume, we can also significantly improve the efficiency by replacing some of them with 2D BEV convolutions. As a result, we achieve superior detection accuracy with more than 4$\times$ speedup in model inference time.

%% file: 4_method.tex
%!TEX root=main.tex
\section{Method}
Recent stereo-based 3D detection models have achieved remarkable progress. The top performing stereo models~\cite{garg2020wasserstein,li2020confidence,pseudoLiDAR,you2019pseudo,qian2020end} follow a similar pipeline, where depth is estimated from stereo features and then  the depth/features are used as points/voxels to perform 3D object detection.
To estimate the depth, all these methods construct cost volumes in the image space, with  disparity as the third dimension.
However, depth reasoning in image space may result in sub-optimal  performance in 3D space tasks.
Moreover, disparity-based cost volumes require   expensive  computations or rely on downsampling which results in loss of accuracy, as they are directly parameterized in terms of the image size. 
To overcome these issues, we propose an efficient stereo-based detector by replacing the disparity cost volume in image space with a novel \textit{p}seudo \textit{L}iDAR feature vol\textit{ume} (\textit{PLUME}) in 3D space. 

Our model consists of four components: (1) A 2D convolutional network  extracts multi-scale features from stereo images; (2) A pseudo LiDAR feature volume is constructed in 3D space and filled with stereo features; (3) A hybrid 3D-BEV convolutional network  performs 3D reasoning; and (4) Multi-task headers  predict 3D occupancy grids as well as object bounding boxes. 
Note that while the proposed PLUME is constructed in 3D space, we output object estimates in bird's eye view (BEV) space, as this is what matters in the downstream tasks of self-driving  (\eg, motion forecasting, motion planning). This is achieved by appending a BEV based detection header. Note however, that the occupancy estimation is performed in 3D space so that the intermediate representations preserve full 3D information. In the case of outputting 3D bounding boxes, we can simply replace the BEV detection header with a 3D based one.
We refer the reader to Fig.~\ref{fig:overview} for an illustration of our model. 

\begin{figure}[t]
\begin{center}
	\includegraphics[width=1.0\linewidth]{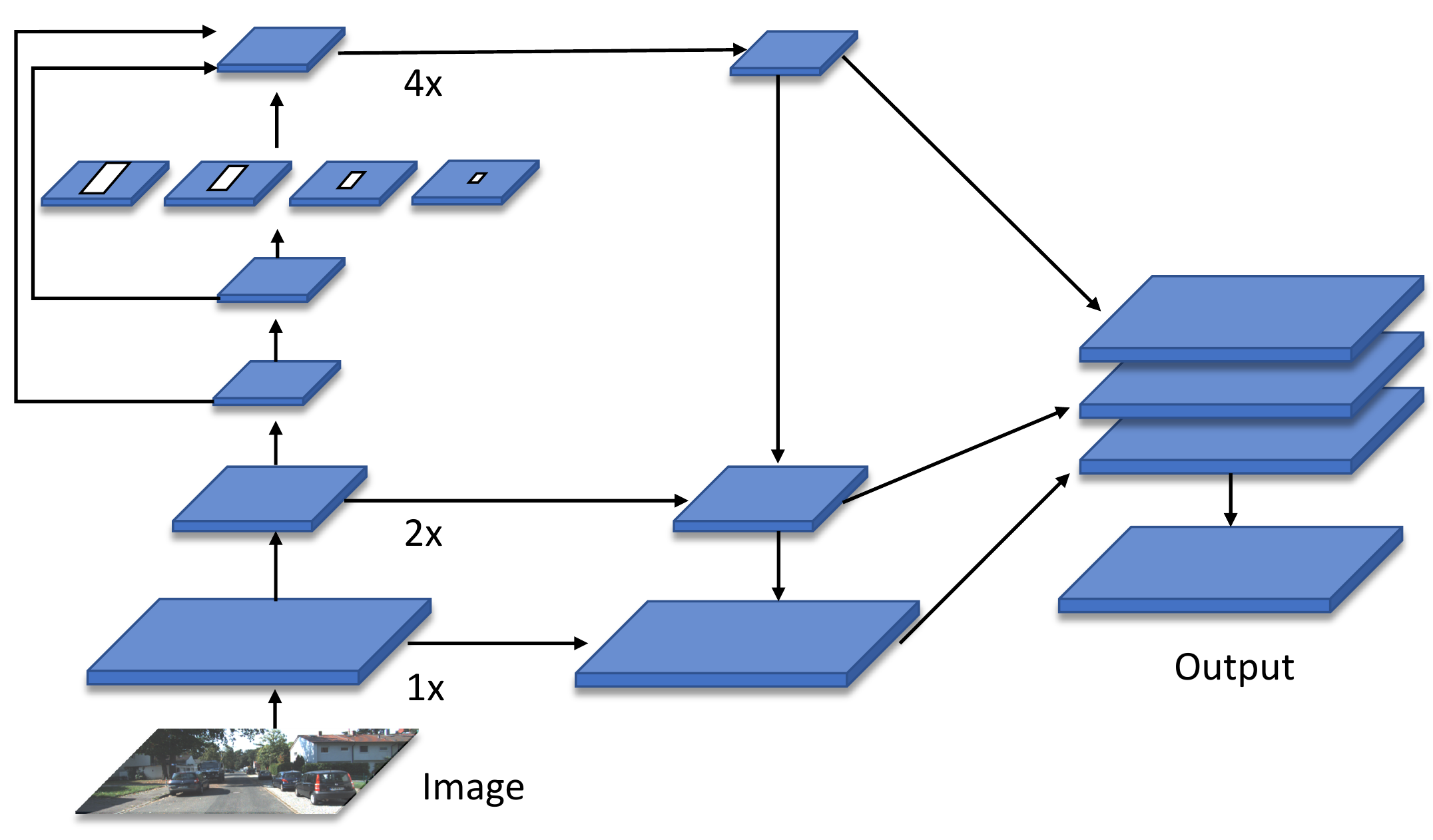}
\end{center}
\vspace{-3mm}
\caption{The stereo image network architecture.}
\label{fig:2dnet}
\end{figure}
\begin{figure}[t]
\begin{center}
	\includegraphics[width=0.7\linewidth]{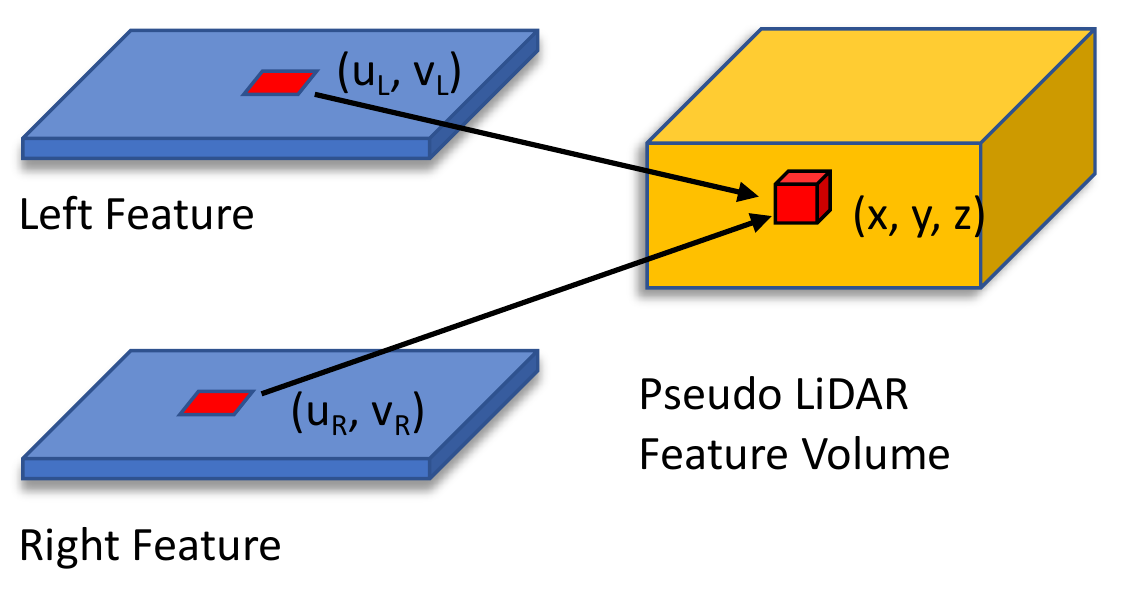}
\end{center}
\vspace{-3mm}
\caption{Illustration of building the pseudo LiDAR feature volume (PLUME). \textcolor{orange}{\textbf{Yellow box}} indicates the 3D feature volume. \textcolor{blue}{\textbf{Blue box}} is 2D feature maps.}
\label{fig:plume}
\end{figure}

\subsection{Stereo Image Network}
We use a 2D convolutional network to extract powerful stereo image features for downstream depth estimation and 3D detection. Stereo images provide disparity cues that disambiguate the depth estimation task when compared to  utilizing monocular images. 
Disparity here refers to the offset of two locations (obtained by projecting a 3D point onto stereo images) along the epipolar line. Since the stereo images are rectified, this search is only horizontal.
Depth is inversely proportional to disparity and can be readily computed from it. As the precision of disparity directly depends on the image resolution, we need high-resolution image features with multi-scale context to achieve good performance.
Based on this consideration we design our stereo image network. Our model uses a spatial pyramid pooling network (SPPN)~\cite{chang2018pyramid} and a feature pyramid network (FPN)~\cite{lin2017feature}  to extract multi-scale features. Importantly, the finest scale in the FPN has the same resolution as the original image. We refer the reader to Fig.~\ref{fig:2dnet} for an illustration of the architecture. 

Note that in other stereo models it is non trivial to use full resolution features. To estimate disparity, these models construct cost volumes by comparing all the stereo pixel pairs with the same vertical coordinate, where the comparison can be implemented as the correlation of two feature vectors, or as a non-linear function parameterized with neural networks (our case).
The cost volume has three dimensions, image height, image width and disparity. If the original image size is used, the 3D cost volume will consume huge amounts of memory and result in costly computation. In contrast, we do not construct a  disparity  volume  and our image feature maps are in 2D, which enable us to use full image resolution efficiently.

\subsection{Pseudo LiDAR Feature Volume (PLUME)}
To perform depth estimation and 3D detection in the same 3D metric space, we directly construct a pseudo LiDAR feature volume. An illustration of PLUME is given in Fig.~\ref{fig:plume}. Specifically, we divide the 3D space in a given range into regular 3D voxels. We assume all voxel centers can be projected onto the images, and use the projection to retrieve the stereo image features generated by our stereo image network. The feature of each voxel is then the concatenation of retrieved image features and voxel coordinates. 
With the constructed feature volume, we apply additional neural network layers to classify each voxel as occupied (object surface) or not. In this way, we are performing an additional  form of depth estimation. However, different from image centric cost volumes in other models, our depth estimation directly operates in 3D space. In the next section, we will describe an efficient neural architecture for PLUME.

\subsection{Efficient 3D-BEV Network}
\begin{figure}[t]
\begin{center}
	\includegraphics[width=\linewidth]{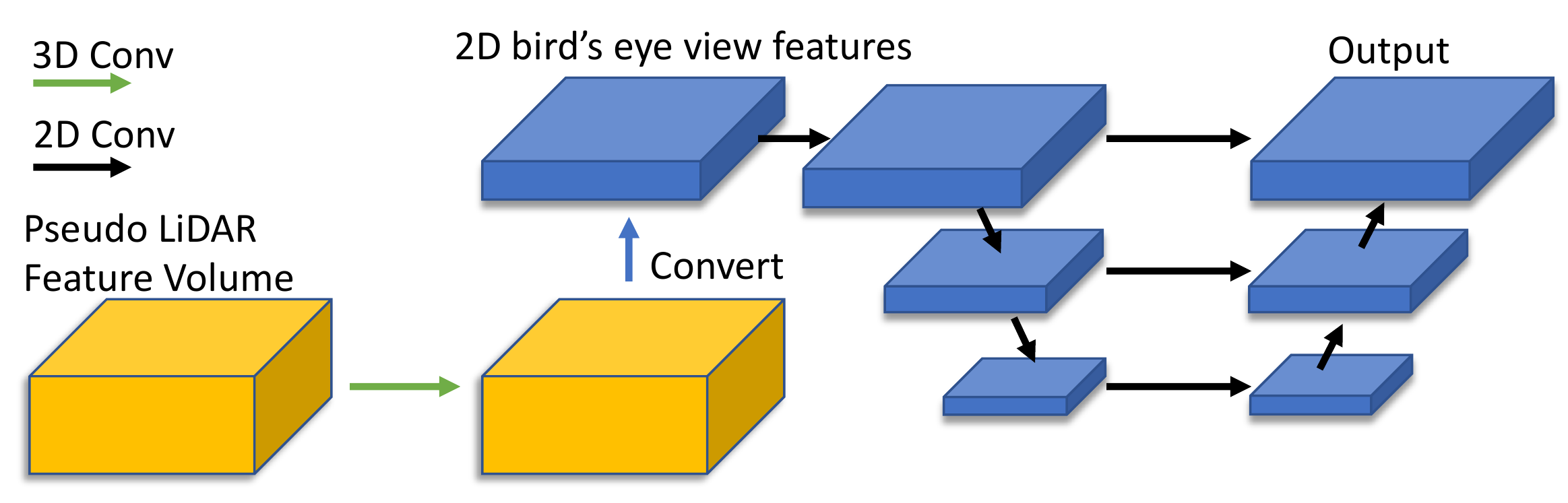}
\end{center}
\vspace{-3mm}
\caption{The 3D-BEV network architecture. \textcolor{orange}{\textbf{Yellow box}} indicates the 3D feature volume. \textcolor{blue}{\textbf{Blue box}} is 2D feature maps.}
\label{fig:2dbev}
\end{figure}

3D convolution is a common choice to extract features from 3D feature volumes. However, compared to 2D convolution, 3D convolution is much more costly in terms of memory and computation. For self-driving perception, the height dimension  is very compact when compared to the other dimensions. For example, the maximum vehicle height rarely exceeds 5 meters, but the detection range in the other two dimensions can easily exceed 80 meters. Motivated by this fact, we propose an efficient 3D-BEV network by replacing 3D convolutions with 2D BEV convolutions. Fig.~\ref{fig:2dbev} shows the overall architecture of the network, which consists of two 3D convolutional layers followed by a 2D hourglass convolutional network. The first two 3D convolutional layers  reason about the scene and object shapes in the height dimension as well as provide a large enough receptive field. 
We then convert the 3D feature volume to a 2D BEV feature map without loss of information by flattening both height and feature dimensions. A 2D hourglass convolutional network then further increases the receptive field in the horizontal and depth dimensions. In Section~\ref{sec:abl}, we show that this hybrid design greatly improves the efficiency while achieving superior performance.

\subsection{Multi-Task Headers}
\paragraph{Occupancy header} The occupancy header consists of two convolutional layers. It takes the output features of our 3D-BEV network as input, and outputs a $D\times W$ 2D feature map with $H$ channels, where the channel dimension represents the height. We reshape this feature map to 3D occupancy voxels, and apply a sigmoid operation to generate the occupancy probability for each voxel.

\paragraph{Detection Header} 
We use an efficient single-stage object detection model based on PIXOR~\cite{yang2018pixor} as our detection header. This model takes the output feature map of our 3D-BEV network as input, and then predicts the  bounding boxes in BEV. It consists of a feature encoder, a feature decoder and a prediction header. The encoder has five blocks of layers.  The first block has two convolutional layers with 32 channels and stride 1. The remaining blocks are composed of bottleneck residual layers~\cite{he2016deep} with number of layers 3, 6, 6, 3 respectively. The output dimension is 16$\times$ smaller than the input. The decoder is an FPN~\cite{lin2017feature} with top-down pathway and lateral connections, outputting feature maps with a 4$\times$ down-sampling factor. The prediction header then takes the FPN fused feature maps and passes them through classification and regression layers to estimate the bounding box size $(w,h)$, BEV location $(u,v)$ and orientation $\theta$ at each pixel.

\begin{table}[t]
\begin{center}
\caption{\textbf{3D object detection results on the KITTI test set.} We report \APBEV of the \textbf{car} category, sorted  at moderate level. All the models take as input only stereo images.}
	\begin{tabular}{@{}l|ccc|c@{}}
		\toprule
		Method & easy & \textit{moderate} & hard & Time (ms) \\\hline
		\SRCNN~\cite{li2019stereo} & 61.9 & 41.3 & 33.4  & 300 \\
		PL~\cite{pseudoLiDAR}    & 67.3  & 45.0 & 38.4  & 400 \\
		RT3DStereo~\cite{konigshof2019realtime} &58.8	&	46.8	&38.4&\textbf{80} \\
		OC-Stereo~\cite{pon2019object} & 68.9 & 51.5 & 43.0  & 350\\
		Disp-RCNN~\cite{sun2020disp} & 73.9 & 52.4 & 43.7 & 420 \\ 
		ZoomNet~\cite{xu2020zoomnet} & 72.9 & 54.9 & 44.1 & 300 \\ 
		PL++~\cite{you2019pseudo}    & 78.3 & 58.0  & 51.3 & 400\\	
		PL-E2E~\cite{qian2020end}   & 79.6 & 58.8 & 52.1 & 400\\ 
		DSGN~\cite{chen2020dsgn} & 82.9 & 65.1 & 56.6 & 670 \\
		CDN~\cite{garg2020wasserstein} & \textbf{83.3} & 66.2 & \textbf{57.7}  & 600\\  \hline
		PLUMENet-Middle   &83.0 &	\textbf{66.3} &	56.7  & 150   \\ \toprule
	\end{tabular}	
	\label{tbl:test}
\end{center}
\end{table}

\subsection{Learning} \label{sec:learning}
We adopt a stage-wise multi-task learning approach. Specifically, we first train the network backbone (stereo image network and 3D-BEV network) plus the occupancy header with depth estimation loss, and then fix the backbone's weights and train the detection header with object detection loss.

We use binary cross entropy (BCE) to compute the occupancy prediction loss. Let $p\in(0,1)$ be the prediction and $y \in \{0, 1\}$ be the ground-truth occupancy of a voxel, where label 1 indicates that the voxel contains at least one LiDAR point.  For each voxel, we define the {\it depth loss} as
\begin{equation}
\sL_{\text{depth}}(p, y)=(y\log(p)+(1-y)\log(1-p))
\end{equation}
Note that we ignore all voxels outside the camera field of view.

The detection loss consists of focal loss for classification and  $\text{smooth}_{\ell_1}$ loss over the bounding box regression terms: size, position and orientation:
\begin{align}
\sL_{\text{detection}}&=\sL_{\text{focal\_loss}}+\sL_{\text{smooth}_{\ell_1}}\\
\sL_{\text{focal\_loss}}(p, c)&=\begin{cases}
-\alpha(1-p)^{\gamma}\log(p) &  \text{if}\; c=1 \\
-(1-\alpha)p^{\gamma}\log(1-p) & \text{otherwise,}
\end{cases}\\
\sL_{\text{smooth}_{\ell_1}}(x)&=\begin{cases}
0.5|x| &  \text{if}\; |x|<0.5 \\
|x|-0.5 & \text{otherwise,}
\end{cases}
\end{align}
where $p, c$ and $ x$ are the class prediction, class label, and distance between predicted and ground-truth regression terms respectively. 

%% file: 5_exp.tex
%!TEX root=main.tex
\section{Experiments}

\begin{figure}[t]
	\centering
		\includegraphics[width=1.0\linewidth]{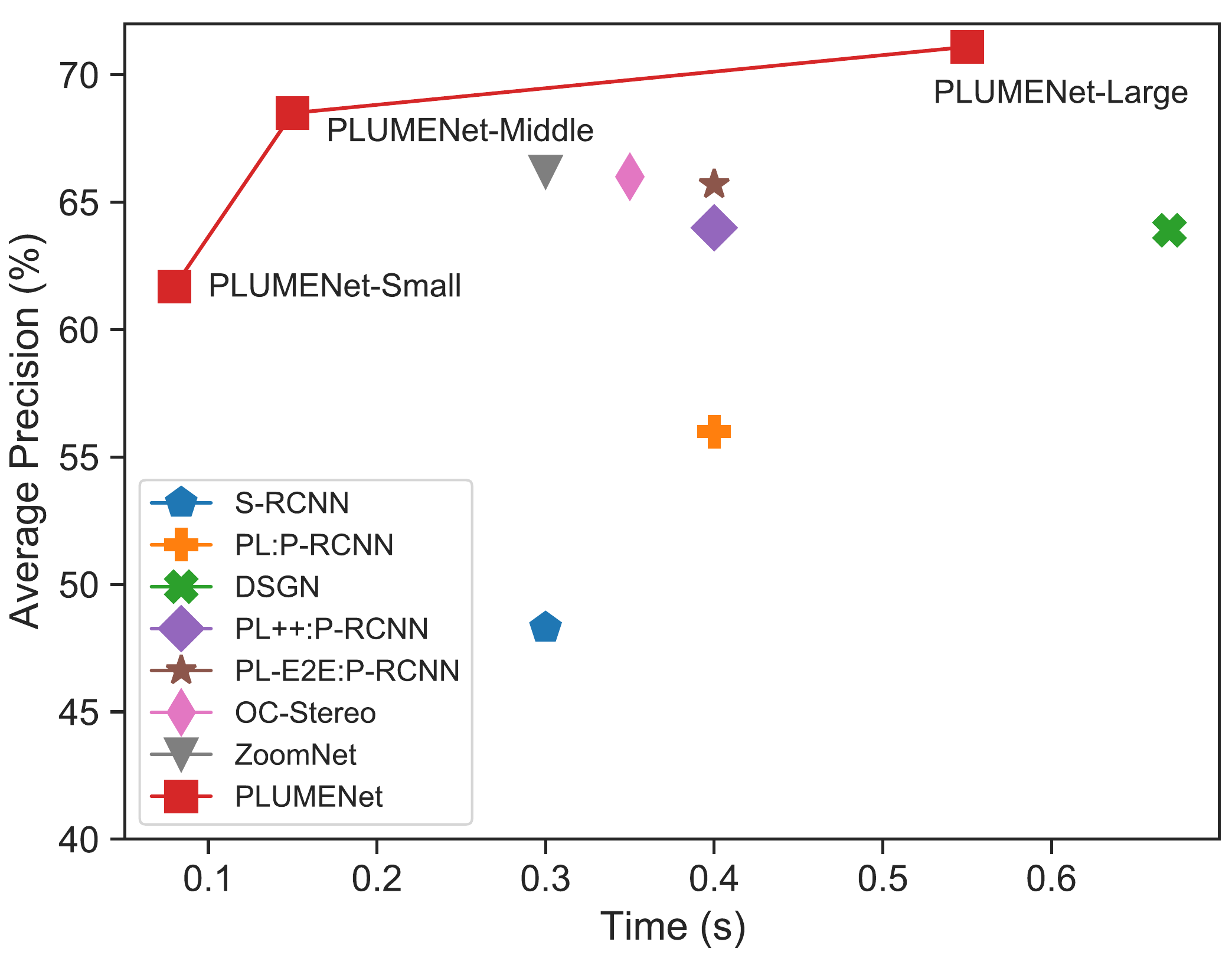}
\vspace{-3mm}
\caption{\textbf{Inference Time v.s. Average Precision} on KITTI~\cite{geiger2013vision} validation set. We report the average precision (AP) in bird's eye view (BEV) at moderate level. Our PLUMENet significantly outperforms other stereo-based object detectors. Specifically, PLUMENet-Middle achieves new state-of-the-art 68.4\% AP while being 2.3$\times$ faster than ZoomNet~\cite{xu2020zoomnet}. PLUMENet-Large achieves 71.1\% AP with a similar speed as DSGN~\cite{chen2020dsgn}.}
\label{fig:time}
\end{figure}

\begin{table*}[t]
\begin{center}
\caption{\textbf{BEV object detection results on the KITTI validation set.} We report \APBEV of the \textbf{car} category. All methods take stereo images as input.}
	\begin{tabular}{l|ccc|ccc|c}
		\toprule
		\multirow{2}{*}{Method} & \multicolumn{3}{c|}{IoU$\ge$0.5}                & \multicolumn{3}{c|}{IoU$\ge$0.7} & \multirow{2}{*}{Time (ms)} \\
		& easy & \textit{moderate} & hard & easy   & \textit{moderate}   & hard   & \\ \hline
		3DOP   \cite{chen20153d}   & 55.0   &   41.3   &  34.6   &  12.6 &  9.5  & 7.6 & -    \\
		MLF   \cite{xu2018multi}   & -   &   53.7   &  -   & -  &  19.5  & - & -    \\
		S-RCNN \cite{li2019stereo} & 87.1  & 74.1   & 58.9   & 68.5  & 48.3  &  41.5 & 300 \\
		PL   \cite{pseudoLiDAR}     & 88.4 & 76.6   &  69.0  & 73.4 &  56.0  & 52.7 & 400    \\
		DSGN     \cite{chen2020dsgn} & -&-& - & 83.2 &  63.9  &57.8  & 670   \\ 
		PL++   \cite{you2019pseudo} & 89.8  & 83.8 & 77.5 & 82.0 & 64.0 & 57.3 & 400    \\
		Disp-RCNN~\cite{sun2020disp}& 90.7 &80.5 &71.0& 77.6 &64.4& 50.7&420\\
		PL-E2E   \cite{qian2020end} &  90.5 & 84.4  &  78.4 & 82.7  &65.7  & 58.4 & 400 \\ 
		OC-Stereo \cite{pon2019object} & 90.0  & 80.6  & 71.1  & 77.7  & 66.0  & 51.2 & 350 \\
		ZoomNet \cite{xu2020zoomnet} &90.6& \textbf{88.4}& 71.4& 78.7 &66.2& 57.6& 300 \\ \hline
		PLUMENet-Small    &87.8& 80.7 &75.2& 74.4& 61.7& 55.8& \textbf{80}     \\
		PLUMENet-Middle   &91.0 &85.9& 80.5 &  83.5& 68.5 &62.8 & 150       \\ 
		PLUMENet-Large   & \textbf{91.3}& 86.6 &\textbf{81.6} &\textbf{84.7}&\textbf{71.1}& \textbf{65.1}& 530        \\ \toprule
	\end{tabular}
\label{tbl:main}
\end{center}
\end{table*}

\begin{table*}[t]
\begin{center}
\caption{\textbf{Ablation results on the KITTI validation set.} We report \APBEV of the \textbf{car} category.}
\begin{tabular}{@{}llcccc@{}}
	\toprule
	\multirow{2}{*}{Experiments}               & \multirow{2}{*}{Method} & \multicolumn{3}{c}{IoU$\ge$0.7} & \multirow{2}{*}{Time (ms)} \\ \cmidrule(lr){3-5}
	&                         & easy    & \textit{moderate}   & hard   &                            \\ \midrule
	\multirow{3}{*}{Image Feature Resolution}        & {\ul Full size}         & \textbf{83.5} & \textbf{68.5} & \textbf{62.8}   & 150                        \\
	& Half Size               & 78.3    & 66.2       & 60.0   & 140                        \\
	& Quarter Size            & 78.2    & 62.1       & 56.1   & \textbf{138}                        \\ \midrule
	\multirow{3}{*}{Feature Volume Network} & {\ul 3D-BEV Network}      & \textbf{83.5} & \textbf{68.5} & 62.8   & 150                        \\
	& BEV Network               & 75.6    & 62.6       & 56.3   & \textbf{110}                        \\
	& 3D Network                 & 82.2    & 68.2       & \textbf{64.8}   & 230                        \\ \midrule
	\multirow{2}{*}{Image Feature Fusion} & {\ul w/o fusion} & 83.5 & 68.5 & 62.8 & \textbf{150} \\
	& with fusion &  \textbf{83.9} & \textbf{68.9} & \textbf{63.4} & 155 \\ \bottomrule 
\end{tabular}
\end{center}
\label{tbl:ablation}
\end{table*}

In this section we first compare PLUMENet 
with  state-of-the-art stereo-based BEV detection models. The results show that our approach significantly outperforms other models, while achieving 2.3$\times$ faster inference speed. We then conduct ablation studies to validate the improvements brought by full resolution image features, the hybrid 3D-BEV network and the pseudo LiDAR feature volume (PLUME).
Finally, we show qualitative results of the proposed model.

\subsection{Experimental Setting}
\paragraph{Dataset}
We evaluate stereo-based BEV object detection on the KITTI detection benchmark~\cite{geiger2013vision}, which contains 7,481 training  and 7,518 test samples. We follow the same training and validation splits used by~\cite{chen20153d}, which divides the training set to 3,712 training  and 3,769 validation samples respectively. 

\paragraph{Evaluation Metric}
We focus on BEV object detection and report results on both the validation and test set. We follow  prior works~\cite{chen2020dsgn, pseudoLiDAR, you2019pseudo} and use Average Precision (AP) computed at 0.5 and 0.7 Intersection-Over-Union (IoU) as our evaluation metric. The KITTI benchmark further splits the labels into three levels: easy, moderate and hard according to the 2D bounding box height in the image, occlusion and truncation level.

\paragraph{Training Details} We set the PLUME range to be $[-32, 32]\times [2, 62.8] \times [-1, 2]$ meters and the voxel resolution to be 0.2 meter. 
We adopt a stage-wise multi-task learning approach: we first train the model with depth loss for 50 epochs, and then train with detection loss for another 50 epochs.
In the first stage we use an initial learning rate of 0.001, and decay by 10$\times$ at epochs 35 and 45 respectively. In the second stage we use an initial learning rate of 0.01 and decay at epochs 30 and 45. Our model is trained on 4 RTX5000 GPUs with a mini-batch size of 8.

\subsection{Experimental Results}
We compare PLUMENet with state-of-the-art stereo based models on BEV object detection. The results on the validation and testing sets are summarized in Table~\ref{tbl:main} and Table~\ref{tbl:test} respectively. 
For a fair comparison we only compare with methods that do not exploit external data and labels.
We designed three different versions of PLUMENet, namely, PLUMENet-Small, PLUMENet-Middle and PLUMENet-Large, using different network sizes by varying the number of feature channels throughout the network architecture. We evaluate the middle version on the testing set, and all three versions on the validation set.
On the test set, our model achieves 66.27\% moderate AP at 0.7 IoU, surpassing all previously published stereo based detectors. Importantly, our model has an inference time of 150 ms (measured on a NVIDIA Tesla V100 GPU), which is 4 $\times$ faster than other high-accuracy detectors \cite{garg2020wasserstein,chen2020dsgn}.

On the validation set, PLUMENet-Small is a real-time prediction model, which runs in only 80ms per frame. It has competitive performance with the state-of-the-art and is more than 3.75$\times$ faster. PLUME-Middle only adds 70ms, but outperforms all other models on both validation and test sets. For example, when IoU$\ge$0.7, PLUME-Middle outperforms the efficient ZoomNet~\cite{xu2020zoomnet} 4.8, 2.3 and 5.2 points on easy, moderate and hard levels respectively, while reducing the inference latency by 70\%. On the test set, PLUMENet-Middle is 4$\times$ faster than  comparable baselines CDN~\cite{garg2020wasserstein} and DSGN~\cite{chen2020dsgn}. In summary, PLUMENet-Middle outperforms all baselines~\cite{chen2020dsgn,pseudoLiDAR,pon2019object,qian2020end,you2019pseudo} that use image-based disparity/depth feature volume, while resulting in much faster inference. 
These results suggest that PLUME is an effective and efficient feature volume representation for the task of stereo based 3D detection.
A detailed detection AP and inference time comparisons on the validation and test sets are summarized in Figs.~\ref{fig:time} and~\ref{fig:time_test}.

\begin{figure*}[ht]
\begin{center}
	\includegraphics[width=1.0\linewidth]{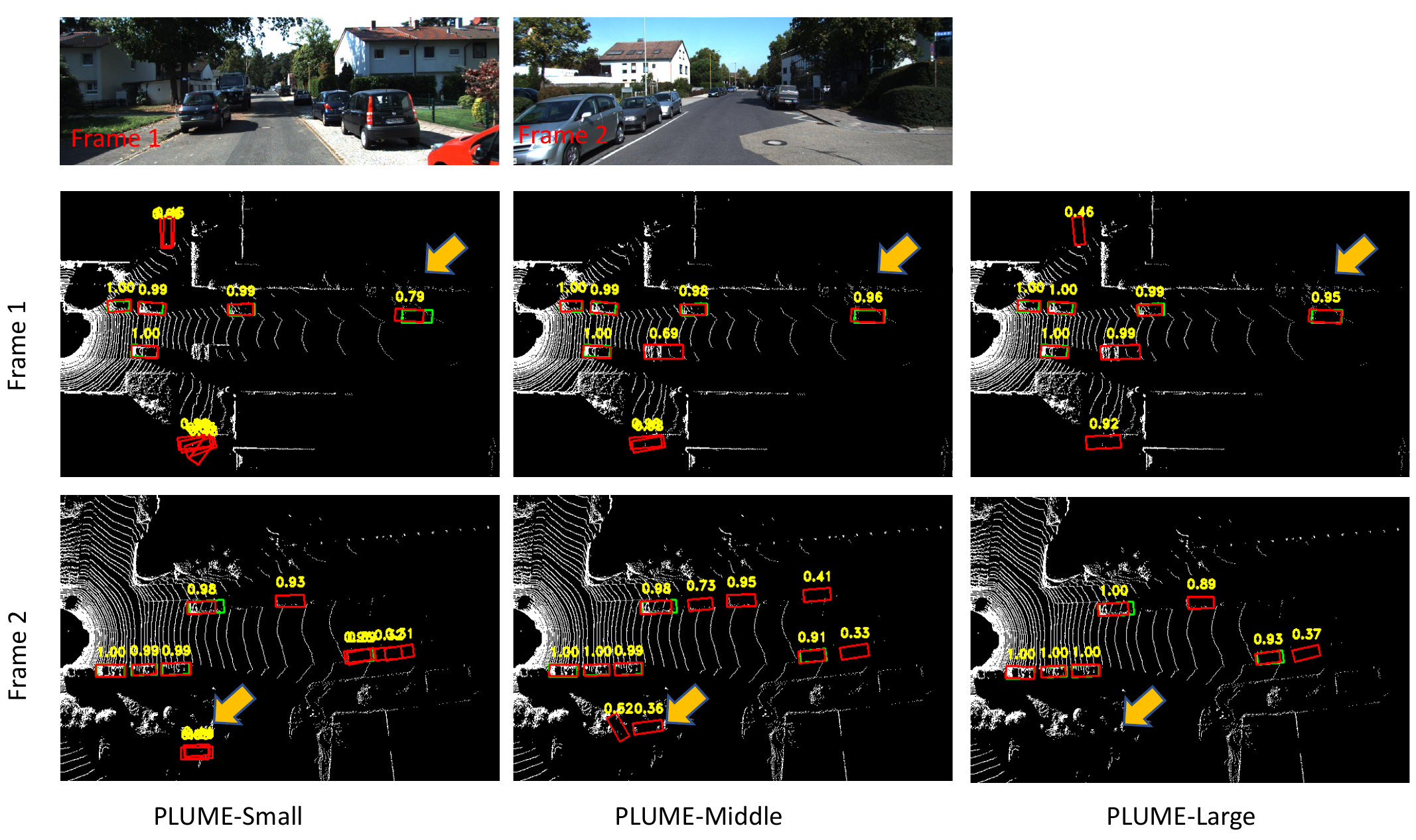}
\end{center}
\vspace{-3mm}
\caption{Qualitative results of our PLUMENet-Small, PLUMENet-Middle and PLUMENet-Large predictions on the KITTI validation set.}
\label{fig:vis}
\end{figure*}

\begin{figure}[t]
\begin{center}
	\includegraphics[width=1.0\linewidth]{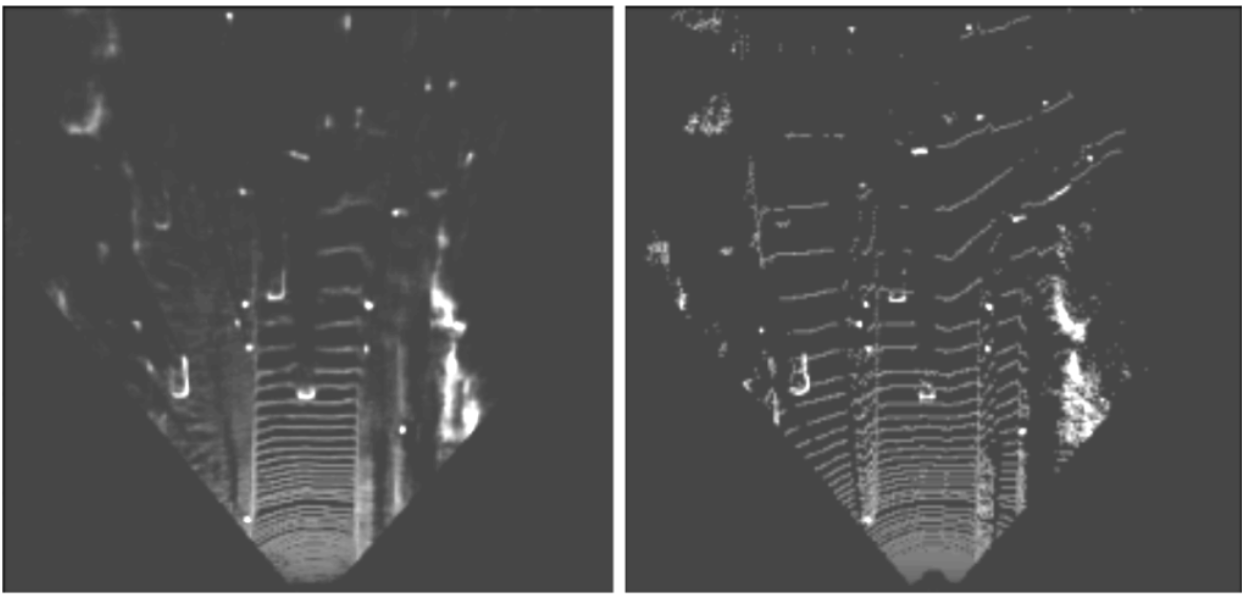}
\end{center}
\vspace{-3mm}
\caption{Predicted occupancy (left) v.s. LiDAR ground-truth (right) in BEV.}
\label{fig:voxel_vis}

\end{figure}

\subsection{Ablation Study}
\label{sec:abl}
We conduct a set of ablation studies to validate the importance of each model component. All these ablations are trained on the KITTI training set and evaluated on the validation set. We use the same learning hyper-parameters for all models. The results are shown in Table~\ref{tbl:ablation}. We use underline to indicate the final model used to compare with other methods.
In each section of the result table, we only vary one component and keep the others unchanged. 

\paragraph{Image Feature Resolution} By default we use the full resolution features extracted from the stereo image network. To achieve the half and quarter-sized resolution (feature map size) while minimizing the impact on  performance, we keep the network encoder and decoder components intact, only adjusting the up-sampling part shown in the last column of Fig.~\ref{fig:2dnet}. The AP gradually drops with the smaller image feature size.

\paragraph{Feature Volume Network} After we build the pseudo LiDAR feature volume from full resolution stereo image feature maps, we apply a feature volume network to process the feature volume and learn discriminative representations for both tasks. We now compare different network architectures for the feature volume network. The BEV network architecture remains similar to the proposed 3D-BEV network, except that the first two 3D convolutional layers are removed. The 3D network replaces all 2D convolutions in 3D-BEV network with 3D ones. Our proposed 3D-BEV network is 6 points better than the pure BEV network while only costing 40 ms extra in runtime. It has competitive performance when compared with a pure 3D network, with a 80ms reduction in inference time.

\paragraph{Image Feature Fusion} We study whether the image features from stereo image network can help detection. Specifically, we build an image feature volume which has the same dimension as PLUME. For each voxel in the volume, we do bilinear warping to get the corresponding left and right image features, weighted by occupancy prediction. We then sum the image volume over height dimension to get bird's eye view features concatenated with features from 3D-BEV network as the input to the detection header. The results in Table~\ref{tbl:ablation} show that image feature fusion is able to slightly improve the model performance by 0.4 points.

\subsection{Qualitative Results}
We show some qualitative results of our three PLUMENets in Fig.~\ref{fig:vis}. The green box is the ground truth and the red box is prediction.  PLUMENet-Large is able to output high quality predictions with less false positives.   PLUMENet-Middle has more false positives than the large model but fewer false negatives than the small one. More specifically, in the first frame, both PLUMENet-Large and PLUMENet-Middle are able to correctly localize the faraway car shown in the top right corner, but the  PLUMENet-Small prediction has a smaller IoU with the ground truth. In the second  frame, PLUMENet-Middle and PLUMENet-Small have a lot of false positives on the bottom part of the figure. We refer the reader to the supplementary video for more qualitative results on KITTI validation set. Fig.~\ref{fig:voxel_vis}  shows the occupancy predictions of our model. PLUMENet can output high quality voxel predictions, where the objects have correct shapes and precise locations.

%% file: 6_discuss.tex
%!TEX root=main.tex
\section{Conclusion}
We have proposed PLUMENet, a model that utilizes efficient pseudo LiDAR feature volume representation for stereo-based BEV object detection. PLUMENet achieves state-of-the-art results on KITTI benchmark~\cite{geiger2013vision} while being 4$\times$ faster than the baselines. We also propose a real-time model which makes reliable predictions in only 80 ms. Furthermore, we build a large PLUMENet model which significantly outperforms the state-of-the-art. 
An extensive ablation study is conducted along with qualitative comparisons to better understand the proposed method. 
Challenges to address in the future include adapting our approach to low-light and extreme weather conditions.

%% file: 7_supp.tex
%!TEX root=main.tex
\begin{table*}[t]
	\scriptsize
	\caption{PLUMENet architecture. \textit{s} and \textit{d} are stride and dilation. \textit{H} and \textit{W} are image height and width. \textit{X}, \textit{Y} and \textit{Z} are feature volume width, height and depth layer \{1,2,3,4\} are cascades of basic residual layers~\cite{he2016deep}.}
	\begin{tabular}{|c|c|c|c|c|c|c|}
		\hline
		\multicolumn{1}{|l|}{} & \multicolumn{2}{c|}{PLUMENet-Small}                                                                                                                  & \multicolumn{2}{c|}{PLUMENet-Middle}                                                                                                                 & \multicolumn{2}{c|}{PLUMENet-Large}                                                                                                                  \\ \hline
		Layer                  & Setting                                                                                               & Output Dimension                             & Setting                                                                                               & Output Dimension                             & Setting                                                                                               & Output Dimension                             \\ \hline
		Image                  &                                                                                                       & $3\times H\times W$                          &                                                                                                       & $3\times H\times W$                          &                                                                                                       & $3\times H\times W$                          \\ \hline
		\multicolumn{7}{|c|}{{\ul \textbf{Stereo Image Network: Feature Encoder}}}                                                                                                                                                                                                                                                                                                                                                                                                                                     \\ \hline
		conv0                  & $(3\times3)\times3$                                                                                   & $32\times H\times W$                         & $(3\times3)\times3$                                                                                   & $32\times H\times W$                         & $(3\times3)\times3$                                                                                   & $32\times H\times W$                         \\ \hline
		maxpool0               & $(3\times3)\times1$, s=2                                                                              & $8\times \tfrac{1}{2}H\times\tfrac{1}{2}W$   & $(3\times3)\times1$, s=2                                                                              & $32\times \tfrac{1}{2}H\times\tfrac{1}{2}W$  & $(3\times3)\times1$, s=2                                                                              & $32\times \tfrac{1}{2}H\times\tfrac{1}{2}W$  \\ \hline
		layer1                 & $(3\times3)\times3$                                                                                   & $8\times \tfrac{1}{2}H\times\tfrac{1}{2}W$   & $(3\times3)\times3$                                                                                   & $32\times \tfrac{1}{2}H\times\tfrac{1}{2}W$  & $(3\times3)\times3$                                                                                   & $32\times \tfrac{1}{2}H\times\tfrac{1}{2}W$  \\ \hline
		layer2                 & $(3\times3)\times6$, s=2                                                                              & $16\times \tfrac{1}{4}H\times\tfrac{1}{4}W$  & $(3\times3)\times6$, s=2                                                                              & $64\times \tfrac{1}{4}H\times\tfrac{1}{4}W$  & $(3\times3)\times6$, s=2                                                                              & $64\times \tfrac{1}{4}H\times\tfrac{1}{4}W$  \\ \hline
		layer3                 & $(3\times3)\times2$                                                                                   & $32\times \tfrac{1}{4}H\times\tfrac{1}{4}W$  & $(3\times3)\times2$                                                                                   & $128\times \tfrac{1}{4}H\times\tfrac{1}{4}W$ & $(3\times3)\times6$                                                                                   & $128\times \tfrac{1}{4}H\times\tfrac{1}{4}W$ \\ \hline
		layer4                 & $(3\times3)\times2$, d=2                                                                              & $32\times \tfrac{1}{4}H\times\tfrac{1}{4}W$  & $(3\times3)\times2$, d=2                                                                              & $128\times \tfrac{1}{4}H\times\tfrac{1}{4}W$ & $(3\times3)\times6$, d=2                                                                              & $128\times \tfrac{1}{4}H\times\tfrac{1}{4}W$ \\ \hline
		branch1                & \begin{tabular}[c]{@{}c@{}}$(64\times64)$ avgpool\\ $(3\times3)$ conv \\ upsample\end{tabular}        & $8\times \tfrac{1}{4}H\times\tfrac{1}{4}W$   & \begin{tabular}[c]{@{}c@{}}$(64\times64)$ avgpool\\ $(3\times3)$ conv \\ upsample\end{tabular}        & $32\times \tfrac{1}{4}H\times\tfrac{1}{4}W$  & \begin{tabular}[c]{@{}c@{}}$(64\times64)$ avgpool\\ $(3\times3)$ conv \\ upsample\end{tabular}        & $32\times \tfrac{1}{4}H\times\tfrac{1}{4}W$  \\ \hline
		branch2                & \begin{tabular}[c]{@{}c@{}}$(32\times32)$ avgpool\\ $(3\times3)$ conv \\ upsample\end{tabular}        & $8\times \tfrac{1}{4}H\times\tfrac{1}{4}W$   & \begin{tabular}[c]{@{}c@{}}$(32\times32)$ avgpool\\ $(3\times3)$ conv \\ upsample\end{tabular}        & $32\times \tfrac{1}{4}H\times\tfrac{1}{4}W$  & \begin{tabular}[c]{@{}c@{}}$(32\times32)$ avgpool\\ $(3\times3)$ conv \\ upsample\end{tabular}        & $32\times \tfrac{1}{4}H\times\tfrac{1}{4}W$  \\ \hline
		branch3                & \begin{tabular}[c]{@{}c@{}}$(16\times16)$ avgpool\\ $(3\times3)$ conv \\ upsample\end{tabular}        & $8\times \tfrac{1}{4}H\times\tfrac{1}{4}W$   & \begin{tabular}[c]{@{}c@{}}$(16\times16)$ avgpool\\ $(3\times3)$ conv \\ upsample\end{tabular}        & $32\times \tfrac{1}{4}H\times\tfrac{1}{4}W$  & \begin{tabular}[c]{@{}c@{}}$(16\times16)$ avgpool\\ $(3\times3)$ conv \\ upsample\end{tabular}        & $32\times \tfrac{1}{4}H\times\tfrac{1}{4}W$  \\ \hline
		branch4                & \begin{tabular}[c]{@{}c@{}}$(8\times8)$ avgpool\\ $(3\times3)$ conv \\ upsample\end{tabular}          & $8\times \tfrac{1}{4}H\times\tfrac{1}{4}W$   & \begin{tabular}[c]{@{}c@{}}$(8\times8)$ avgpool\\ $(3\times3)$ conv \\ upsample\end{tabular}          & $32\times \tfrac{1}{4}H\times\tfrac{1}{4}W$  & \begin{tabular}[c]{@{}c@{}}$(8\times8)$ avgpool\\ $(3\times3)$ conv \\ upsample\end{tabular}          & $32\times \tfrac{1}{4}H\times\tfrac{1}{4}W$  \\ \hline
		\multicolumn{2}{|c|}{concat(layer\{2,4\}, branch\{1,2,3,4\})}                                                                  & $80\times \tfrac{1}{4}H\times\tfrac{1}{4}W$  &                                                                                                       & $320\times \tfrac{1}{4}H\times\tfrac{1}{4}W$ &                                                                                                       & $320\times \tfrac{1}{4}H\times\tfrac{1}{4}W$ \\ \hline
		conv1                  & $(3\times3)\times1$                                                                                   & $32\times \tfrac{1}{4}H\times\tfrac{1}{4}W$  & $(3\times3)\times1$                                                                                   & $128\times \tfrac{1}{4}H\times\tfrac{1}{4}W$ & $(3\times3)\times1$                                                                                   & $128\times \tfrac{1}{4}H\times\tfrac{1}{4}W$ \\ \hline
		conv2                  & $(1\times1)\times1$                                                                                   & $8\times \tfrac{1}{4}H\times\tfrac{1}{4}W$   & $(1\times1)\times1$                                                                                   & $32\times \tfrac{1}{4}H\times\tfrac{1}{4}W$  & $(1\times1)\times1$                                                                                   & $32\times \tfrac{1}{4}H\times\tfrac{1}{4}W$  \\ \hline
		\multicolumn{7}{|c|}{{\ul \textbf{Stereo Image Network: Feature Decoder}}}                                                                                                                                                                                                                                                                                                                                                                                                                                \\ \hline
		fpn\_conv2             & \begin{tabular}[c]{@{}c@{}}input: conv1\\ $(1\times1)\times1$\end{tabular}                            & $32\times \tfrac{1}{4}H\times\tfrac{1}{4}W$  & \begin{tabular}[c]{@{}c@{}}input: conv1\\ $(1\times1)\times1$\end{tabular}                            & $64\times \tfrac{1}{4}H\times\tfrac{1}{4}W$  & \begin{tabular}[c]{@{}c@{}}input: conv1\\ $(1\times1)\times1$\end{tabular}                            & $96\times \tfrac{1}{4}H\times\tfrac{1}{4}W$  \\ \hline
		fpn\_conv2\_up         & $\begin{bmatrix}(3\times3) \text{conv}\\\text{upsample}\end{bmatrix}\times2$                          & $32\times H\times W$                         & $\begin{bmatrix}(3\times3) \text{conv}\\\text{upsample}\end{bmatrix}\times2$                          & $64\times H\times W$                         & $\begin{bmatrix}(3\times3) \text{conv}\\\text{upsample}\end{bmatrix}\times2$                          & $96\times H\times W$                         \\ \hline
		fpn\_conv1             & \begin{tabular}[c]{@{}c@{}}input: layer1\\ $(1\times1)$ conv\\ upsample fpn\_conv2\\ sum\end{tabular} & $32\times \tfrac{1}{2}H\times\tfrac{1}{2}W$  & \begin{tabular}[c]{@{}c@{}}input: layer1\\ $(1\times1)$ conv\\ upsample fpn\_conv2\\ sum\end{tabular} & $64\times \tfrac{1}{2}H\times\tfrac{1}{2}W$  & \begin{tabular}[c]{@{}c@{}}input: layer1\\ $(1\times1)$ conv\\ upsample fpn\_conv2\\ sum\end{tabular} & $96\times \tfrac{1}{2}H\times\tfrac{1}{2}W$  \\ \hline
		fpn\_conv1\_up         & $\begin{bmatrix}(3\times3) \text{conv}\\\text{upsample}\end{bmatrix}\times1$                          & $32\times H\times W$                         & $\begin{bmatrix}(3\times3) \text{conv}\\\text{upsample}\end{bmatrix}\times1$                          & $64\times H\times W$                         & $\begin{bmatrix}(3\times3) \text{conv}\\\text{upsample}\end{bmatrix}\times1$                          & $96\times H\times W$                         \\ \hline
		fpn\_conv0             & \begin{tabular}[c]{@{}c@{}}input: conv0\\ $(1\times1)$ conv\\ upsample fpn\_conv1\\ sum\end{tabular}  & $32\times H\times W$                         & \begin{tabular}[c]{@{}c@{}}input: conv0\\ $(1\times1)$ conv\\ upsample fpn\_conv1\\ sum\end{tabular}  & $64\times H\times W$                         & \begin{tabular}[c]{@{}c@{}}input: conv0\\ $(1\times1)$ conv\\ upsample fpn\_conv1\\ sum\end{tabular}  & $96\times H\times W$                         \\ \hline
		fpn\_conv0\_up         & $(3\times3)\times1$                                                                                   & $32\times H\times W$                         & $(3\times3)\times1$                                                                                   & $64\times H\times W$                         & $(3\times3)\times1$                                                                                   & $96\times H\times W$                         \\ \hline
		\multicolumn{2}{|c|}{sum(fpn\_conv\{0,1,2\}\_up)}                                                                              & $32\times H\times W$                         &                                                                                                       & $64\times H\times W$                         &                                                                                                       & $96\times H\times W$                         \\ \hline
		dropout                & p=0.2                                                                                                 & $32\times H\times W$                         & p=0.2                                                                                                 & $64\times H\times W$                         & p=0.2                                                                                                 & $96\times H\times W$                         \\ \hline
		fpn\_conv              & $(3\times3)\times1$                                                                                   & $32\times H\times W$                         & $(3\times3)\times1$                                                                                   & $32\times H\times W$                         & $(3\times3)\times1$                                                                                   & $32\times H\times W$                         \\ \hline
		\multicolumn{7}{|c|}{{\ul \textbf{3D-BEV Network}}}                                                                                                                                                                                                                                                                                                                                                                                                                                         \\ \hline
		\multicolumn{2}{|c|}{Pseudo-LiDAR Feature Volume}                                                                              & $64\times X\times Y \times Z$                &                                                                                                       & $64\times X\times Y \times Z$                &                                                                                                       & $64\times X\times Y \times Z$                \\ \hline
		3dconv0                & $(3\times3\times3)\times2$                                                                            & $12\times X\times Y \times Z$                & $(3\times3\times3)\times2$                                                                            & $32\times X\times Y \times Z$                & $(3\times3\times3)\times2$                                                                            & $48\times X\times Y \times Z$                \\ \hline
		\multicolumn{2}{|c|}{reshape 3D feature volume to 2D}                                                                          & $(12\times Y)\times X \times Z$              &                                                                                                       & $(32\times Y)\times X \times Z$              &                                                                                                       & $(48\times Y)\times X \times Z$              \\ \hline
		bev\_conv0             & $(3\times3)\times2$                                                                                   & $96\times X \times Z$                        & $(3\times3)\times2$                                                                                   & $160\times X \times Z$                       & $(3\times3)\times2$                                                                                   & $256\times X \times Z$                       \\ \hline
		bev\_conv1             & \begin{tabular}[c]{@{}c@{}}$(3\times3)\times2$\\ add bev\_conv0\end{tabular}                          & $96\times X \times Z$                        & \begin{tabular}[c]{@{}c@{}}$(3\times3)\times2$\\ add bev\_conv0\end{tabular}                          & $160\times X \times Z$                       & \begin{tabular}[c]{@{}c@{}}$(3\times3)\times2$\\ add bev\_conv0\end{tabular}                          & $256\times X \times Z$                       \\ \hline
		bev\_conv2             & $(3\times3)\times2$, s=2                                                                              & $192\times \tfrac{1}{2}X\times\tfrac{1}{2}Z$ & $(3\times3)\times2$, s=2                                                                              & $320\times \tfrac{1}{2}X\times\tfrac{1}{2}Z$ & $(3\times3)\times2$, s=2                                                                              & $512\times \tfrac{1}{2}X\times\tfrac{1}{2}Z$ \\ \hline
		bev\_conv3             & $(3\times3)\times2$, s=2                                                                              & $192\times \tfrac{1}{4}X\times\tfrac{1}{4}Z$ & $(3\times3)\times2$, s=2                                                                              & $320\times \tfrac{1}{4}X\times\tfrac{1}{4}Z$ & $(3\times3)\times2$, s=2                                                                              & $512\times \tfrac{1}{4}X\times\tfrac{1}{4}Z$ \\ \hline
		bev\_deconv4           & \begin{tabular}[c]{@{}c@{}}$(3\times3)\times2$ deconv\\ add bev\_conv2\end{tabular}                   & $192\times \tfrac{1}{2}X\times\tfrac{1}{2}Z$ & \begin{tabular}[c]{@{}c@{}}$(3\times3)\times2$ deconv\\ add bev\_conv2\end{tabular}                   & $320\times \tfrac{1}{2}X\times\tfrac{1}{2}Z$ & \begin{tabular}[c]{@{}c@{}}$(3\times3)\times2$ deconv\\ add bev\_conv2\end{tabular}                   & $512\times \tfrac{1}{2}X\times\tfrac{1}{2}Z$ \\ \hline
		bev\_deconv5           & $(3\times3)\times2$ deconv                                                                            & $96\times X \times Z$                        & $(3\times3)\times2$ deconv                                                                            & $160\times X \times Z$                       & $(3\times3)\times2$ deconv                                                                            & $256\times X \times Z$                       \\ \hline
		\multicolumn{7}{|c|}{{\ul \textbf{Occupancy Header}}}                                                                                                                                                                                                                                                                                                                                                                                                                                       \\ \hline
		conv3                  & $(3\times3)\times1$                                                                                   & $48\times X \times Z$                        & $(3\times3)\times1$                                                                                   & $80\times X \times Z$                        & $(3\times3)\times1$                                                                                   & $128\times X \times Z$                       \\ \hline
		conv4                  & $(3\times3)\times1$                                                                                   & $Y \times X \times Z$                        & $(3\times3)\times1$                                                                                   & $Y \times X \times Z$                        & $(3\times3)\times1$                                                                                   & $Y \times X \times Z$                        \\ \hline
	\end{tabular}
\label{tbl:plm}
\end{table*}

\section{Appendix: Model Architecture}
\label{arch}
We provide PLUMENet small, middle and large architectures in Table~\ref{tbl:plm}, which consist of stereo image network, 3D-BEV network and occupancy header. Both the occupancy header and detection header take as input the bev\_conv5 in Table~\ref{tbl:plm}.  The detection header consists of five blocks of layers. We show the number of layers and the output channels of each block in Table~\ref{tbl:resconf}. We refer readers to PIXOR~\cite{yang2018pixor} for more details of the detection header. 

\begin{table}[t]
	\centering
	\caption{Configurations of the detection header.}
	\label{tbl:resconf}
	\begin{tabular}{lcc}
		\toprule
		& number of layers & number of channels     \\ \hline
		Small  & 2, 3, 6, 6, 3    & 32, 96, 128, 192, 192  \\ \hline
		Middle & 2, 3, 6, 6, 3    & 32, 96, 192, 256, 384  \\ \hline
		Large  & 3, 3, 6, 6, 3    & 48, 128, 192, 256, 384 \\ \bottomrule
	\end{tabular}
\end{table}